\documentclass{article}

\PassOptionsToPackage{numbers}{natbib}
\usepackage[final]{nips_2017}

\usepackage[utf8]{inputenc} 
\usepackage[T1]{fontenc}    
\usepackage{hyperref}       
\usepackage{url}            
\usepackage{booktabs}       
\usepackage{amsfonts}       
\usepackage{nicefrac}       
\usepackage{microtype}      
\usepackage{times}
\usepackage{url}
\usepackage{amsmath,amssymb}
\usepackage{amsfonts}
\usepackage{graphicx}
\usepackage{float}
\usepackage{subcaption}
\usepackage{comment}
\usepackage{color}

\definecolor{deepmagenta}{rgb}{0.8, 0.0, 0.8}

\title{DLTK: State of the Art Reference Implementations for Deep Learning on Medical Images} 
\author{Nick Pawlowski \And Sofia Ira Ktena \And Matthew C.H. Lee \AND Bernhard Kainz \And Daniel Rueckert \And Ben Glocker \And Martin Rajchl \AND
Biomedical Image Analysis Group\\Department of Computing, Imperial College London, UK\\
\texttt{np716@imperial.ac.uk}
}

\begin{document}

\maketitle

\begin{abstract}
We present DLTK, a toolkit providing baseline implementations for efficient experimentation with deep learning methods on biomedical images. It builds on top of TensorFlow and its high modularity and easy-to-use examples allow for a low-threshold access to state-of-the-art implementations for typical medical imaging problems. A comparison of DLTK's reference implementations of popular network architectures for image segmentation demonstrates new top performance on the publicly available challenge data ``Multi-Atlas Labeling Beyond the Cranial Vault''. The average test Dice similarity coefficient of $81.5$ exceeds the previously best performing CNN ($75.7$) and the accuracy of the challenge winning method ($79.0$).
\end{abstract}

\section{Introduction}
The successful application of deep convolutional neural networks (CNNs) to the ImageNet challenge \cite{Russakovsky2015} by Krizhevsky et al. \cite{Krizhevsky2012} has had a large impact on the field of computer vision. Biomedical image analysis is a particular area benefiting from latest advancements in computer vision methodology, but it only recently adapted deep learning techniques on a wider scale \cite{Litjens2017}.
Adaptation of deep learning to biomedical imaging problems requires speciality operations and the lack of low-threshold access to validated reference implementations has led to slow progress. To address this, we present DLTK\footnote{\url{https://dltk.github.io/}}, a Deep Learning Toolkit for Medical Image Analysis. DLTK simplifies the application and experimentation with deep learning methods on medical imaging data by providing validated, high-performance reference implementations and required speciality operations.

\section{Building DLTK}
We believe that most of the deep learning research relevant to medical imaging will address at least one of these components: a) data reading and preprocessing; b) model definitions and network architectures; c) training and optimisation strategies; d) deployment of methods to new data.
Therefore, DLTK enables a plug-and-play structure for those components based on TensorFlow's \cite{Abadi2015} new high-level API \cite{cheng2017tensorflow}. This approach gives users the freedom to reuse existing components but also to rapidly integrate new functionality as subject of their research. We naturally interface within the TensorFlow framework and therefore benefit from its wide range of operations and community contributions. While other packages, such as NiftyNet \cite{niftynet17}, build on a predefined application structure, we emphasise an API level prioritising experimentation by exposing access to low-level operations. 

\section{Experiments \& Results}
DLTK reference implementations of FCN \cite{Long2015} and U-Net \cite{Ronneberger2015} architectures using residual units \cite{He2015} are tested on the publicly available challenge dataset from the MICCAI 2015 challenge ``Multi-Atlas Labeling Beyond the Cranial Vault''. We test combinations of components, \textit{a)} data reading with random and class-balanced sampling of patches, \textit{b)} training with a Dice, cross-entropy and class-balanced cross-entropy loss for the two network architectures. All methods are trained with ADAM \cite{Kingma2014} with default parameters and tuned $\hat{\epsilon}=10^{-5}$ to counteract loss spikes. Network inputs are patches of size $64^{3}$ voxels. We found that the U-Net trained with cross-entropy loss and class-balanced sampling performed best out of all combinations. The comparative performance of all experiment dimensions is depicted in Figure \ref{fig:network_box_plot} in the Appendix.

For external comparison, we submitted the best performing method to the challenge website and compare to the winning entry employing a multi-atlas approach \cite{heinrichmulti} and the best performing CNN \cite{Larsson2017}. Our U-Net implementation achieves new state-of-the-art results of $81.5$ in terms of DSC compared to the multi-atlas method by \cite{heinrichmulti} with $79.0$. The previously best performing CNN \cite{Larsson2017} achieves $75.7$. Interestingly, we report a slightly lower validation performance of $78.9$ compared to \cite{Larsson2017} with $79.0$ which might indicate overfitting of the previous CNN \cite{Larsson2017} to the training data as we achieve better test set performance than validation performance. We further note that our architecture is far from its potential optimal performance as we did not fine-tune any of its hyperparameters or the data preprocessing and rather prefer to report out-of-the-box performance of the DLTK implementation.

\begin{figure*}[h]
\centering
\begin{subfigure}{0.45\textwidth}
	\centering
	\includegraphics[clip,height=6cm]{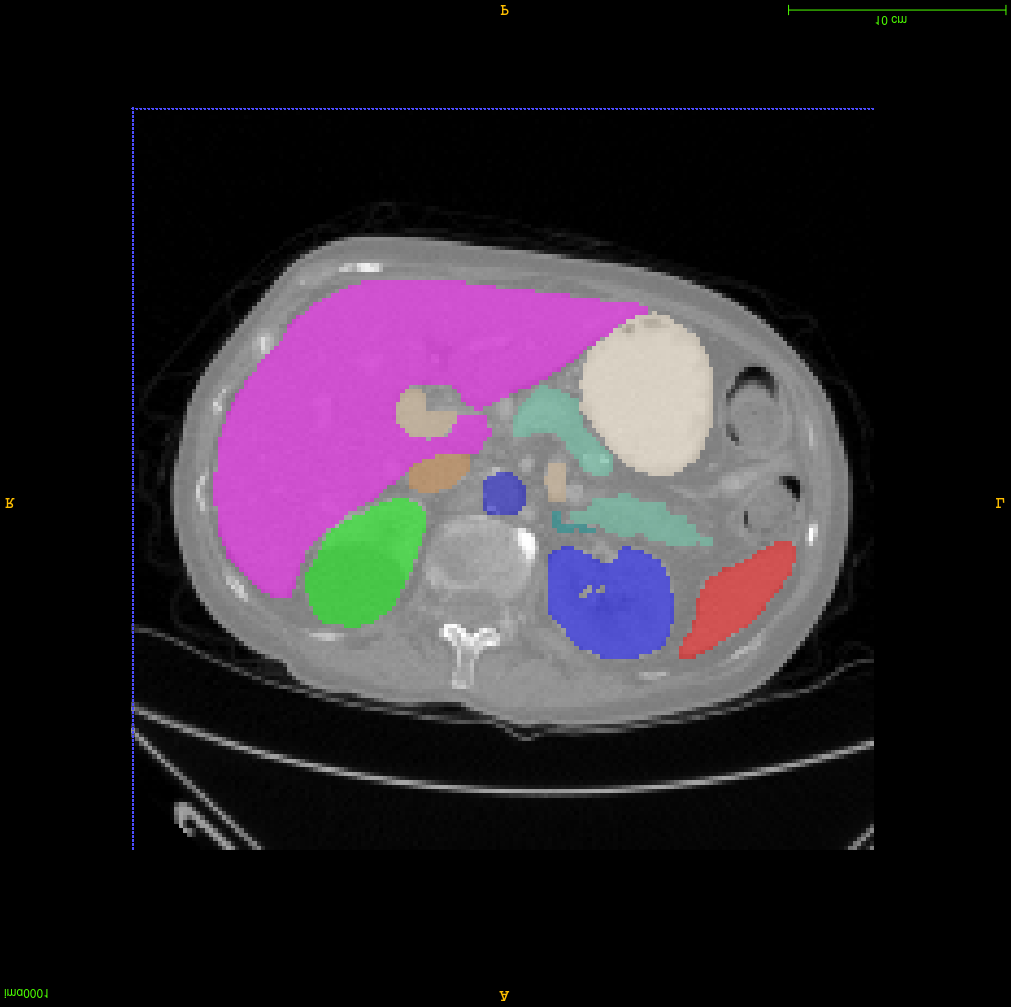}
    \caption{Axial view of an exemplary segmentation}
    \label{fig:ct_ax}
	\end{subfigure}%
    \,
  \begin{subfigure}{0.45\textwidth}
	\centering
	\includegraphics[clip,height=6cm]{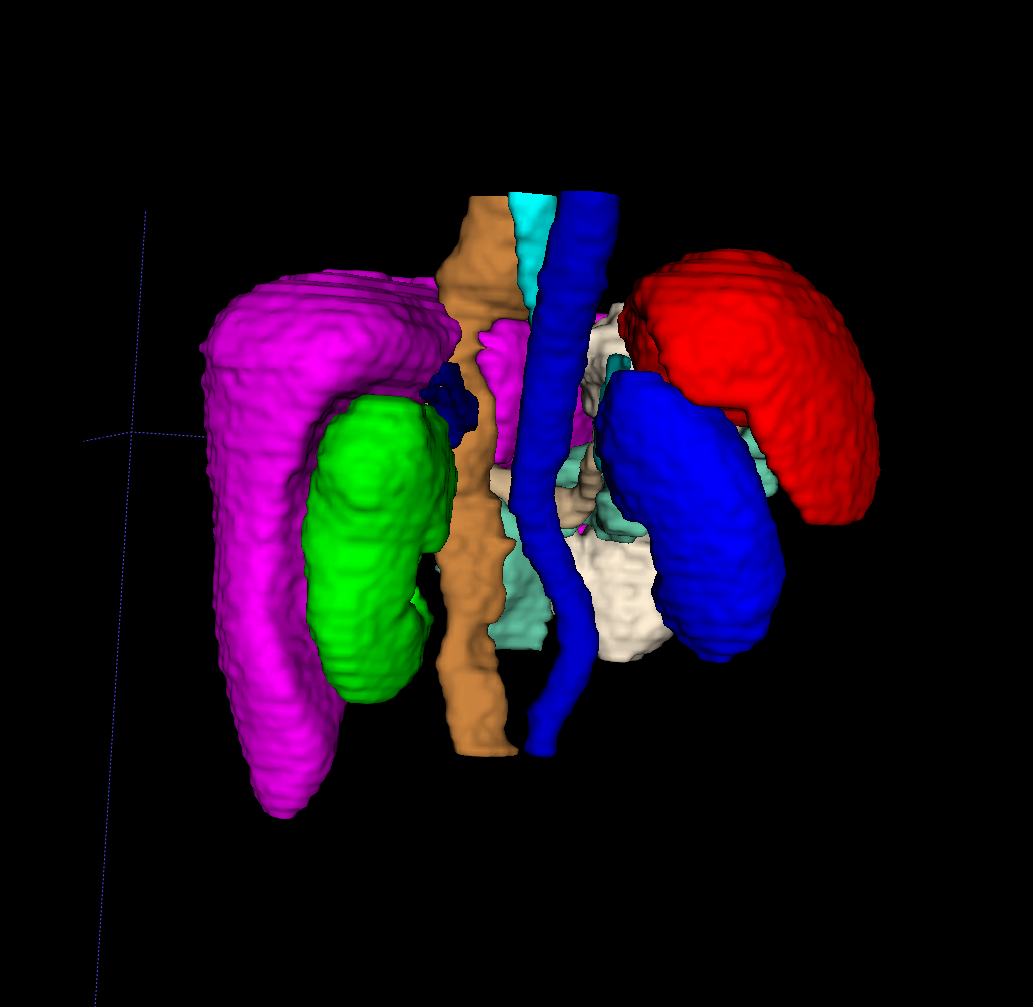}
    \caption{3D rendering of an exemplary segmentation}
    \label{fig:ct_seg}
\end{subfigure}%
\caption{Prediction of the DLTK U-Net segmenting 13 organs on abdominal CT scans.}
\label{fig:segmentation}
\end{figure*}

\section{Conclusion}
We present a new tool kit for performing deep learning experiments tailored to biomedical image analysis. DLTK offers basic baseline implementations for popular network architectures, sampling techniques and losses commonly used in medical image analysis. It further offers easy deployment including sliding-window inference invariant to input shapes. In future, we aim to extend the range of DLTK to include more additional components and latest developments and thus, enable low-threshold access to cutting-edge deep learning methods for medical imaging. 

\section*{Acknowledgements}
NP is supported by Microsoft Research PhD Scholarship and the EPSRC Centre for Doctoral Training in High Performance Embedded and Distributed Systems (HiPEDS, Grant Reference EP/L016796/1). MR is supported by an Imperial College Research Fellowship. We gratefully acknowledge the support of NVIDIA with the donation of one Titan X GPU for our research.

\bibliographystyle{plain}
\bibliography{biblio}

\newpage
\appendix
\section{Experimental Comparison of Network Architectures}

\begin{figure*}[h]
\centering
\includegraphics[width=14cm]{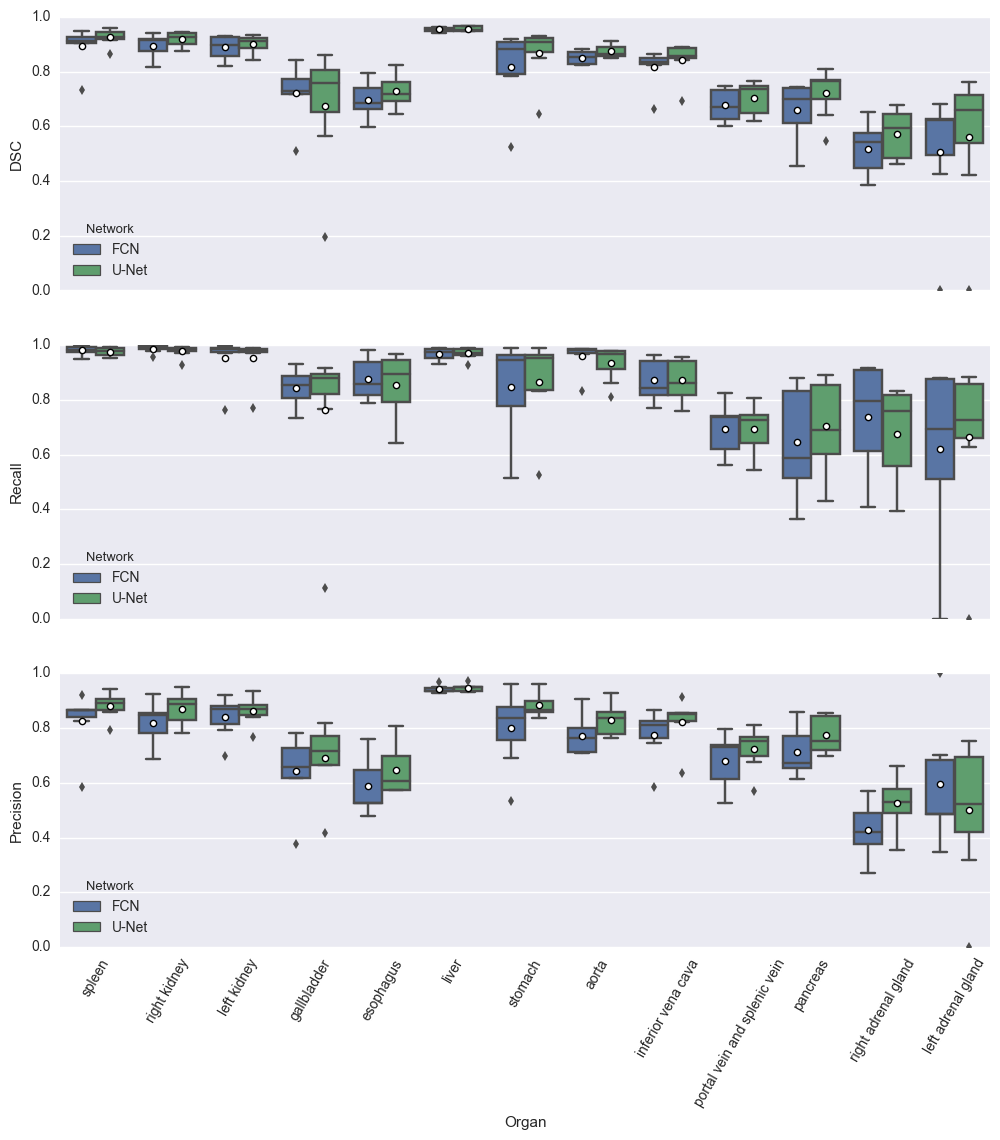}
\caption{Box plot comparing the DSC scores, recall and precision for the U-Net (green) and FCN (blue). The white markers show the mean, the black line shows the median, the error bars indicate confidence intervals, and the additional black markers indicate outliers. The U-Net outperforms the FCN in almost every comparison. The FCN only outperforms the U-Net on the recall for the right adrenal gland.}
\label{fig:network_box_plot}
\end{figure*}

\end{document}